# D-Score: A Synapse-inspired Approach for Filter Pruning

Doyoung Park[1, *], Jinsoo Kim[2, *], Jina Nam[3, 4], Jooyoung Chang[5], and Sang Min Park[6, 7, †]

[1]e0225083@u.nus.edu
[2]jinsookim@yonsei.ac.kr
[3]INTHESMART Co., Ltd., Seoul, Republic of Korea
[4]Department of Clinical Medical Sciences, Seoul National University College of Medicine, Seoul, Republic of Korea
[5]Department of Biomedical Sciences, Seoul National University Graduate School, Seoul, Republic of Korea
[6]Department of Biomedical Sciences, Seoul National University College of Medicine, Seoul, Republic of Korea
[7]XAIMED, Seoul, Republic of Korea

April 4, 2022

## Abstract

This paper introduces a new aspect for determining the rank of the unimportant filters for filter pruning on convolutional neural networks (CNNs). In the human synaptic system, there are two important channels known as excitatory and inhibitory neurotransmitters that transmit a signal from a neuron to a cell. Adopting the neuroscientific perspective, we propose a synapse-inspired filter pruning method, namely Dynamic Score (D-Score). D-Score analyzes the independent importance of positive and negative weights in the filters and ranks the independent importance by assigning scores. Filters having low overall scores, and thus low impact on the accuracy of neural networks are pruned. The experimental results on CIFAR-10 and ImageNet datasets demonstrate the effectiveness of our proposed method by reducing notable amounts of FLOPs and Params without significant Acc. Drop.

## 1 Introduction

Convolutional neural networks (CNNs) have proven their usefulness in many computer vision fields: face recognition, image classification, and human pose estimation [1, 11, 21, 29, 36, 37]. However, to fully utilize their superior performances, high computational power and large memory space are mandatory [6]. Such limitations refrain from deploying CNNs on resource-constrained devices like embedded systems and mobile phones [26, 38]. To solve the problems of CNNs with a large number of parameters and model size, numerous model compression methods which are mainly classified as low-rank factorization [31], knowledge distillation [5, 17], quantization [3, 12], and pruning [13, 22] have been widely studied over the past years. Among the compression methods, pruning is a technique to remove unimportant and redundant parameters from CNNs and has demonstrated its effectiveness by reducing the computational cost and increasing the inference speed [7].

Depending on the type of parameters for elimination, pruning can be renamed as weight pruning [13], neuron pruning [35], and filter pruning [22]. Due to the technical mechanism of weight pruning and neuron pruning, they remove specific weight connections and neurons, and this introduces sparsity in the network. On the contrary, filter pruning discards the entire unnecessary filters from the neural networks and this maintains the structured network that guarantees the compatibility of the filter-pruned networks with existing libraries and hardware [22, 28].

In the current literature on filter pruning, many studies [15, 22, 42] analyze the weight values in the filters to select less important filters for pruning. Recent study by [22] employed $\ell_1$-norm method to select the unimportant filters for filter pruning (Figure 1). However, to the best of our knowledge, no study has separately considered the positive and negative weights in the filters for determining unimportant filters for pruning.

---

*Contributed equally.

†Corresponding author.

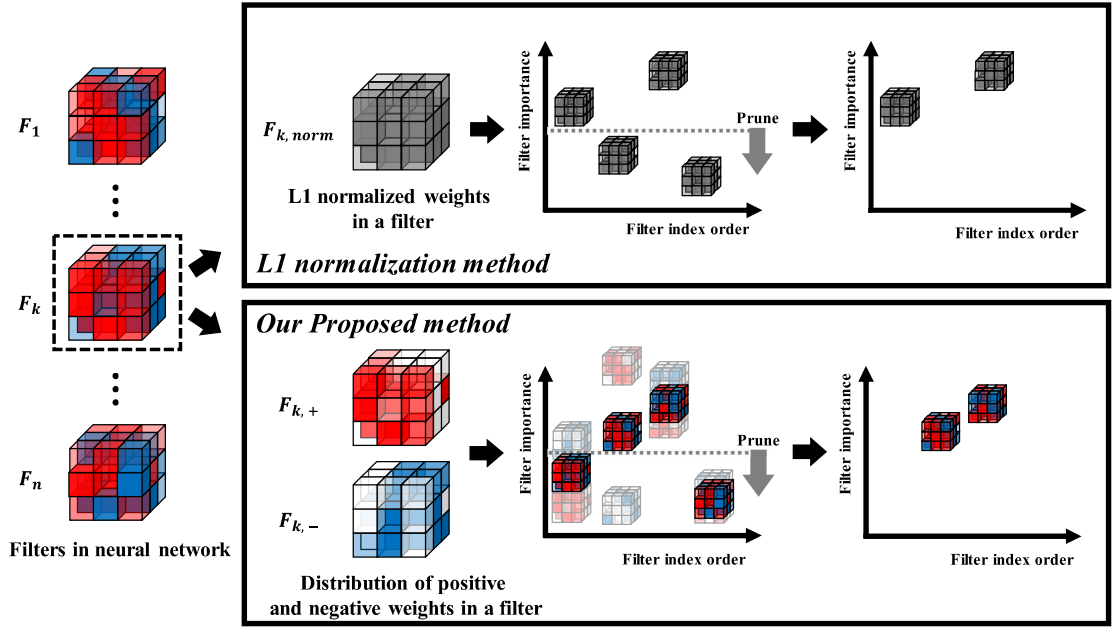

Figure 1: Difference in the concept of $\ell_1$-norm approach and D-Score approach for determining the rank of the filters. Different filters are pruned as the rank of the filters varies depending on the analyzing approaches. For $\ell_1$-norm approach, the rank of the filters is determined by calculating the importance using $\ell_1$-norm of each filter. For D-Score approach, the rank of the filters is determined by calculating the independent importance of positive and negative weights and assigning scores to them.

In the synaptic transmission system of a biological neural network, the basis of an artificial neural network, there coexists excitatory and inhibitory neurotransmitters. The excitatory neurotransmitter enhances or increases the activation in the postsynaptic membrane, while the inhibitory neurotransmitter decreases or prevents the activation [4, 18, 19]. Similarly, filters of CNN models are composed of positive weights and negative weights. Considering the neuroscientific perspective, we propose a new filter pruning approach that separately analyzes the positive and negative weights in the filters, namely Dynamic Score (D-Score). D-Score assigns scores to the filters for filter pruning based on their independent importance of positive and negative weights (Figure 1). In addition to D-Score, two variants called Dynamic Step (D-Step) and Dynamic Step with Geometric Median (D-Step GM) are also introduced as applied concepts of our proposed approach.

Our contributions of this paper are summarized as below:

- We propose a new filter pruning technique called D-Score, that independently processes and assigns scores to the positive and negative weights in the filters based on their independent importance. In addition to D-Score, two more applied concepts, namely D-Step and D-Step GM, are also introduced. All three approaches yield improved results as compared to other proposed filter pruning methods in terms of the amount of reduction in floating points operation per second (FLOPs) and the number of parameters (Params), and accuracy drop (Acc. Drop).

- We prove that the correlation of positive and negative weights when determining the rank of the unimportant filters is crucial.

## 2   Related works

Several studies have stated that one of the essential directions of filter pruning is to develop innovative methods for selecting redundant filters for pruning [7, 24]. As discussed in the introduction, [22] employed $\ell_1$-norm method to select and prune filters having a small effect on the accuracy of the neural network. [40] proposed an iterative filter pruning method that used Taylor expansion as the estimation for determining the rank of the unimportant filters. [15] introduced a soft pruning method which used $\ell_p$-norm method to select the unimportant filters and prune them by adjusting weights as zero. For practical purpose, [15] employed $\ell_2$-norm to select the unimportant filters. [39] introduced a stochastic training method that froze certain channels to be constant for pruning. [16] adopted geometric median to calculate the Euclidean spaces [9] to determine filters for pruning based on their redundancy, not importance. [24] analyzed the corresponding feature maps of the filters and pruned the filters having low-rank feature maps. [42] employed a clustering algorithm called spectral clustering to select a group of filters and evaluate the importance by $\ell_p$-norm for pruning. [33] proposed an adaptive filter pruning method that iteratively pruned the filters when the accuracy drop was within the acceptable range. [28] detected weak channels first and pruned their associated filters in the previous layer.

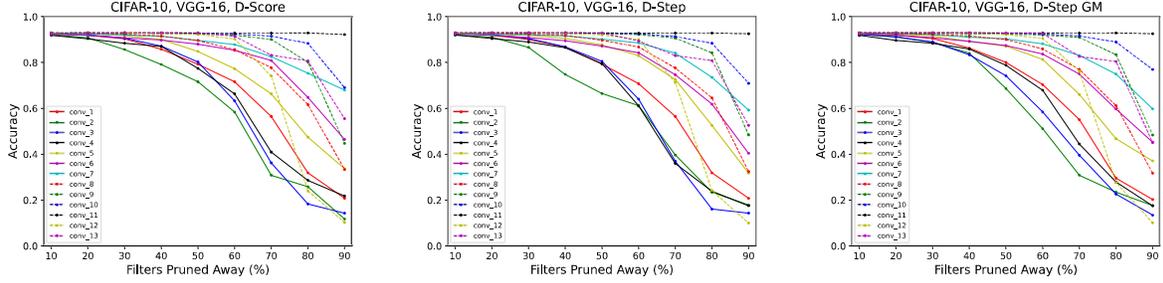

Figure 2: Sensitivity analysis of VGG-16 trained on CIFAR-10 dataset using D-Score, D-Step, and D-Step GM. Different methods produce different sensitivity patterns.

## 3 The Proposed Methods

### 3.1 Independent ranking of positive and negative filters

Let $\mathcal{F}_j^i$ denotes the $j$th filter in $i$th layer, and each filter $\mathcal{F}_j^i$ comprises of $\mathbb{R}^{C_i \times K_i \times K_i}$, where $C_i$ stands for channel, and $K_i$ for width and height of a kernel. $\mathcal{F}_j^i$ is composed of positive and negative weights (Eq. 1),

$$\mathcal{F}_j^i = \mathcal{F}_{j,+}^i + \mathcal{F}_{j,-}^i \tag{1}$$

and the independent sum of the positive weights and negative weights for each filter are denoted as (Eq. 2).

$$\mathcal{F}_{j,+}^i = \sum_{n=1}^N \mathcal{W}_{n,+} \;\; and \;\; \mathcal{F}_{j,-}^i = \sum_{n=1}^N \mathcal{W}_{n,-} \tag{2}$$

**Dynamic Score**    The detailed procedure of D-Score method is as follows:

1. Using the values calculated from Eq. 2, sort the positive filters in ascending order and the negative filters in descending order.

2. Assign scores of $[1, j]$ to the positive and negative filters independently according to their sorted orders.

3. Find the overall score of the filter $\mathcal{F}_j^i$ using the Eq. 1, and sort the filters $\mathcal{F}_j^i$ in ascending order of their overall scores.

4. Prune the filters with small scores based on the sensitivity analysis (Section 3.2).

**Dynamic Step**    The detailed procedure of D-Step method is as follows:

1. Using the values calculated from Eq. 2, sort the positive filters in ascending order and the negative filters in descending order.

2. Set a buffer size equivalent to the pruning threshold (Section 3.3) for step-wise comparison of sorted positive and negative filters.

3. Fill the buffer with filters $\mathcal{F}_j^i$ in which the values of their positive and negative filters are simultaneously positioned close to 0.

4. Prune the filters positioned front in the buffer based on the sensitivity analysis (Section 3.2).

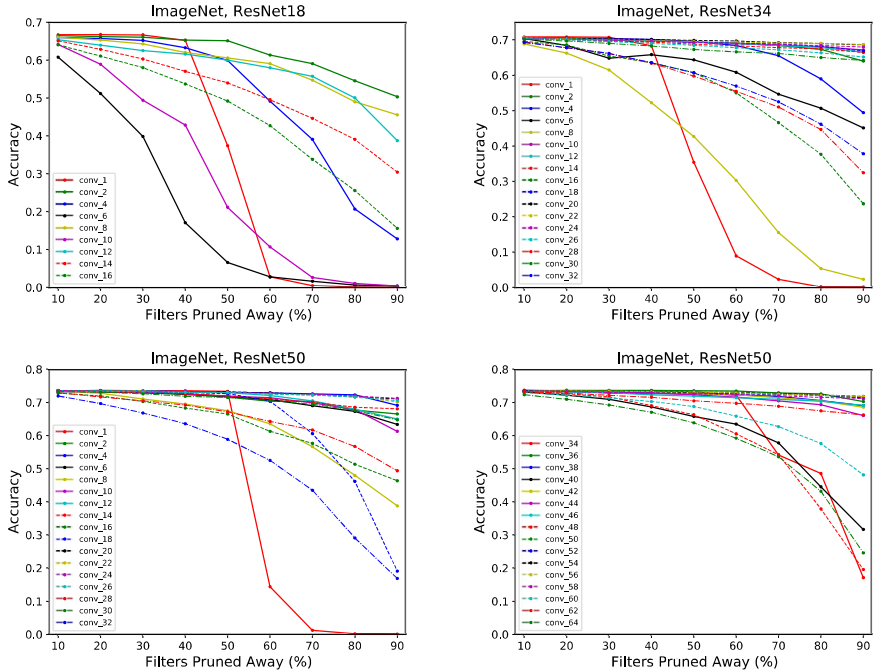

Figure 3: Sensitivity analysis of ResNet18, 34, 50 trained on ImageNet dataset using D-Score.

**Dynamic Step with Geometric Median**    The detailed procedure of D-Step GM method is as follows:

1. Applying the idea discussed in [9, 16], calculate the independent Euclidean distances of positive and negative filters and sort them in the ascending order of their distances.

2. Set a buffer size equivalent to the pruning threshold (Section 3.3) for step-wise comparison of sorted positive and negative filters.

3. Fill the buffer with filters $\mathcal{F}_j^i$ in which the values of their positive and negative filters are simultaneously positioned close to the shortest distance.

4. Prune the filters positioned close to the shortest distance in the buffer based on the sensitivity analysis (Section 3.2).

## 3.2   Pruning Sensitivity Analysis

To determine the sensitivity of individual layers to pruning, we iteratively pruned each layer and evaluated the accuracy of the pruned network in every step [22]. This procedure was repeated for all the proposed methods as shown in Figure 2, 3. Figure 2 shows that different pruning methods produced different sensitivities to pruning in the same layer. Figure 3 shows that the accuracy of layers that are sensitive to pruning decreased significantly as more filters were pruned, and this pattern is clearly noticeable in the first convolutional layer of ResNets.

## 3.3   Parallel Pruning and Retraining Pruned Models

Several studies have demonstrated that pruning convolutional layers decreases computational load and pruning fully connected layers decreases sizes of neural networks [2, 22, 35]. For neural networks comprised of convolutional and fully connected layers such as VGGNet [32], all relevant layers were pruned away based on the pruning sensitivity analysis. For neural networks containing residual blocks such as ResNet [14], we omitted to prune the last convolutional layer in each block due to the specificity of their architectures.

Based on the pruning sensitivity analysis, we set a pruning threshold accuracy (acc) that was used to calculate the number of filters to be eliminated in all applicable layers for parallel pruning. The parallel pruning is a time-saving technique as it prunes the different number of filters in all applicable layers at once. For example, D-Score in Figure 2 shows that setting the pruning threshold accuracy as 0.9 (acc. 90%) pruned 30%, 45%, 90% of filters at once in layer 1, 7, 11 respectively.

Table 1: Comparison of our methods and other pruning methods for VGG-16 trained on CIFAR-10 dataset. For CIFAR-10, top-1 Acc. Drop, Params reduction, and FLOPs reduction are compared. The best performance is in bold.

| Model | Approach | Acc. Drop(%) | Params Reduction(%) | FLOPs Reduction(%) |
|-------|----------|--------------|---------------------|--------------------|
| VGG-16 | PFEC [22] | **-0.15** | 64.0 | 34.2 |
|  | FPGM [16] | 0.04 | – | 34.2 |
|  | NSP [41] | -0.04 | – | 54.0 |
|  | HRank [24] | 0.53 | 82.9 | 53.5 |
|  | NS [27] | -0.14 | **88.52** | 51.0 |
|  | Ours (D-Score) | 0.16 | 87.03 | 64.81 |
|  | Ours (D-Step) | 0.12 | 86.70 | **65.40** |
|  | Ours (D-Step GM) | -0.10 | 87.16 | 64.37 |

Upon completion of parallel pruning of a neural network, the pruned network has poor performance compared to the original model. The performance of the pruned network can be restored close to the original network by retraining for fewer epochs than the one used for training the original model.

# 4 Experiments

We evaluated the performance of the proposed techniques based on the sensitivity analysis with two representative datasets and various network structures: CIFAR-10 dataset [20] with VGG-16 [32], and ImageNet ILSVRC-2012 dataset [30] with ResNet18, 34, and 50 [14]. Since D-Step and D-Step GM are the applied concepts of D-Score, the performance of D-Step and D-Step GM were only experimented with CIFAR-10 dataset. We calculated top-1 accuracy for CIFAR-10 dataset, and top-1 and top-5 accuracy for ImageNet dataset respectively. All three techniques yielded outperforming results.

## 4.1 Experimental Setup

**Initial model for CIFAR-10**   Since CIFAR-10 is a small dataset, we trained VGG-16 from scratch for 450 epochs with a batch size of 64. The initial learning rate was set to 0.01 and decayed by 10% in every 20 epochs of training. During training the baseline model, the data augmentation with horizontal flip, random width and height shifts, and rotation of 15 degrees was used.

**Initial model for ImageNet**   Since ImageNet is a large dataset, we adopted pre-trained models for ResNet18, 34, and 50. Before pruning ResNet models, the pre-trained ResNet18, 34, and 50 were retrained with ImageNet in TFRecord format for 10 epochs with batch sizes of 512, 256, and 128 respectively. During this process, the upper and lower bounds of the learning rates for respective models were derived by the learning rate finder [34], and the data augmentation with horizontal flip, random cropping, random color distortion, and rotation of 15 degrees was used.

## 4.2 Experiment Results

**CIFAR-10**   We compared the performance of our proposed methods with other pruning methods in terms of the Acc. Drop, reduction in Params and FLOPs. Note that we only calculated the top-1 accuracy for CIFAR-10 dataset as it only contains 10 classes, and the negative accuracy drop indicates that the accuracy of the pruned and retrained model is higher than the accuracy of the original model. According to Table 1, all of our proposed methods significantly outperformed other methods on VGG-16. In comparison with [22], another filter importance ranking-based approach, the performance of D-Step GM was substantially better as D-Step GM yielded similar Acc. Drop with significantly higher reduction in both Params and FLOPs. Among our proposed methods, D-Step GM yielded the highest reduction in Params by 87.16% with an increase in the final accuracy by 0.1%. For VGG-16 with CIFAR-10 dataset, the experimental result implies that the independent ranking of the positive and negative weights to select the unimportant filters for pruning led to a high reduction in Params and FLOPs without significant Acc. Drop.

Table 2: Comparison of our method and other filter pruning methods for ResNet18, 34, 50 trained on ImageNet dataset. For ImageNet, top-1 and top-5 Acc. Drop, and FLOPs reduction are compared. The best performance is in bold.

| Model | Approach | Top-1 Acc. Drop(%) | Top-5 Acc. Drop(%) | FLOPs Reduction(%) |
|---|---|---|---|---|
| ResNet18 | PFP [23] | 1.08 | 0.50 | 19.99 |
| | Ours (D-Score) | **0.16** | **0.03** | 23.93 |
| | FPGM [16] | 1.87 | 1.15 | 41.8 |
| | FBS [10] | 2.54 | 1.46 | – |
| | LCCN [8] | 3.65 | 2.30 | 34.6 |
| | SFP [15] | 3.18 | 1.85 | 41.8 |
| | Ours (D-Score) | 1.76 | 0.96 | **49.24** |
| ResNet34 | PFEC [22] | 1.06 | – | 24.2 |
| | Ours (D-Score) | 0.78 | 0.36 | 30.25 |
| | FPGM[16] | 1.29 | 0.54 | 41.1 |
| | LCCN [8] | **0.43** | **0.17** | 24.8 |
| | SFP [15] | 2.09 | 1.29 | 41.1 |
| | Ours (D-Score) | 1.72 | 0.87 | **43.01** |
| ResNet50 | PFP [23] | **0.22** | **0.06** | 10.82 |
| | Ours (D-Score) | 1.23 | 1.01 | 14.78 |
| | ThiNet [28] | 0.84 | 0.47 | 36.7 |
| | SFP [15] | 14.01 | 8.27 | 41.8 |
| | GDFP [25] | 2.52 | 1.25 | 41.97 |
| | FPGM [16] | 1.32 | 0.55 | 53.5 |
| | Ours (D-Score) | 1.99 | 1.25 | **53.64** |

**ImageNet** We compared the performance of D-Score with other pruning methods in terms of Acc. Drop and reduction in FLOPs. In contrast to CIFAR-10, we calculated the top-1 and top-5 accuracy for ImageNet. According to Table 2, when D-Score yielded the highest reduction in FLOPs for ResNet18 (49.24%) and ResNet34 (43.01%), the top-1 Acc. Drop by D-Score in each model was smaller compared to that of the other pruning methods with a similar reduction in FLOPs. When D-Score yielded the highest reduction in FLOPs for ResNet50 (53.64%), it also had higher top-1 and top-5 Acc. Drop than several other methods. This result can be due to the tradeoff between the reduction in FLOPs and recovery for accuracy. For ResNet18, D-Score showed remarkably better performance with the smallest top-1 Acc. Drop (0.16%) and top-5 Acc. Drop (0.03%) with a higher reduction in FLOPs than [23]. For ResNet34, D-Score was substantially better than [22], another filter importance ranking based-approach, by resulting in small Acc. Drop with higher reduction in FLOPs. This result implies that the D-Score was an efficient method to maintain a fine balance between the Acc. Drop and reduction in FLOPs.

## 4.3 Analysis

In this section, we compared one of our proposed methods, D-Score, with another filter importance ranking based-approach [22], that employed $\ell_1$-norm for calculating the filter importance.

**Visualization of Filters and Feature maps** Based on the pruning sensitivity analysis (Figure 2), we pruned 20% of filters using D-Score. Figure 4 reveals that depending on the methods for ranking the importance of filters, their overall ranking varied. Different ranking of filters pruned different combinations of filters and this notably influenced the reduction in FLOPs and Params. According to Table 1, while showing similar Acc. Drop, D-Score yielded 87.03% reduction in Params and 64.81% reduction in FLOPs while $\ell_1$-norm-based approach [22] resulted in 64% reduction in Params and 34.3% reduction in FLOPs. Figure 4 shows that certain corresponding feature maps of pruned filters were replaceable by the remaining feature maps. The feature maps of pruned filters 0 and 14 by D-Score were similar to the feature maps of the remaining filters 7 and 51. On the contrary, $\ell_1$-norm removed filters 4 and 17 which produced unique feature maps.

**Weight Distribution after Pruning** For comparison, we pruned the same number of filters, 20% of filters as discussed in the previous subsection, using both D-Score and $\ell_1$-norm. Figure 5 shows that a neural network pruned by D-Score contained filters composed of either more positive or more negative weights. As discussed in the above subsection, retaining positive-prone and negative-prone filters in CNNs facilitated a higher reduction in Params and FLOPs while

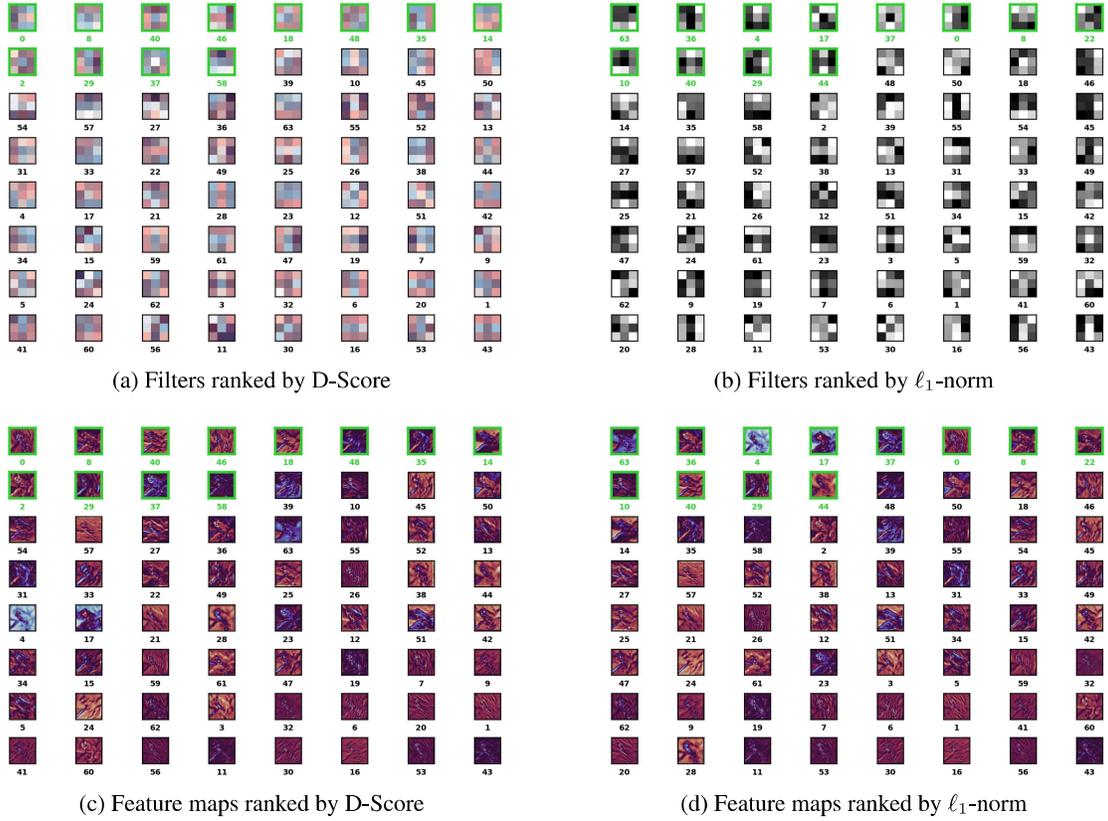

(a) Filters ranked by D-Score

(b) Filters ranked by $\ell_1$-norm

(c) Feature maps ranked by D-Score

(d) Feature maps ranked by $\ell_1$-norm

Figure 4: Visualization of filters and the corresponding feature maps of the second convolutional layer of VGG-16 ranked by D-Score and $\ell_1$-norm. The filters and feature maps are displayed horizontally in the ascending order of their ranks. For filters of D-Score, we color the positive-prone and negative-prone weights red and blue respectively. Filters and feature maps in the green boxes are to be pruned. For D-Score, 20% of filters are to be pruned based on the sensitivity analysis. For comparison, the same number of filters and feature maps to be pruned using $\ell_1$-norm are marked with the green boxes.

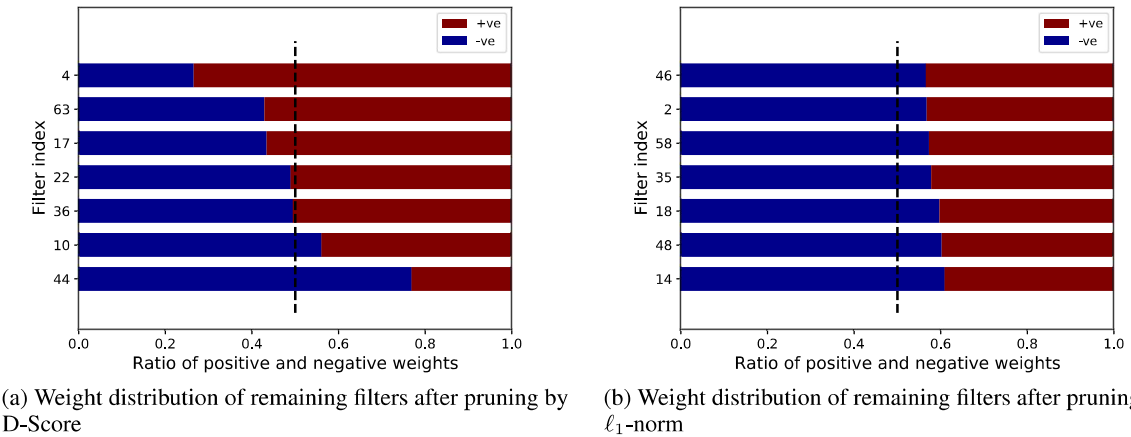

(a) Weight distribution of remaining filters after pruning by D-Score

(b) Weight distribution of remaining filters after pruning by $\ell_1$-norm

Figure 5: Ratio of positive and negative weights in the remaining filters after pruning by D-Score and $\ell_1$-norm. For comparison, only the filters with the different indices among the remaining filters are visualized.

maintaining similar Acc. Drop compared to $\ell_1$-norm-based approach. Higher reduction in Params and FLOPs leads to higher compression rate for model size and faster inference speed. Therefore, it can be deduced that preserving positive-prone or negative-prone filters in CNNs played an important role in reducing more Params and FLOPs without significant Acc. Drop, similar to the importance of the excitatory and inhibitory neurons in the human synaptic system.

## 5 Conclusion

In this paper, we propose a synapse-inspired innovative method for determining the rank of the unimportant filters in CNNs for filter pruning. For neurotransmission in the synapse, both excitatory and inhibitory neurotransmitters, responsible for increasing and decreasing the activation respectively, play a decisive role in signal transmission. Similarly, we selected the unimportant filters for pruning by measuring the independent importance of positive and negative weights in the filter. To the best of our knowledge, this is the first study to consider the correlation of positive and negative weights in the filters when determining the rank of the filters for filter pruning. We showed that neural networks pruned by our method preserved positive-prone or negative-prone filters and this resulted in reducing more Params and FLOPs without significant Acc. Drop. However, our study includes several limitations. First, our study was only conducted with two types of models, VGGNet and ResNet. Second, the applied concepts of D-Score, namely D-Step and D-Step GM were not experimented with ResNet. Therefore, validation of the performance of D-Score and the applied concepts, D-Step and D-Step GM, with other types of models such as MobileNet remains as our future work. Through this study, we demonstrated that the correlation of positive and negative weights in the filters is crucial when determining the unimportant filters for pruning.